\documentclass{easychair}
\usepackage{tipa}
\usepackage{url}
\usepackage[utf8]{inputenc}
\usepackage[T1]{fontenc}

\usepackage{doc}

\title{RIMAX: Ranking Semantic Rhymes\\ by calculating Definition Similarity\thanks{This work was partially financed by Avignon Universit\'e (France) and El Colegio de México (Mexico).}}

\author{
Alfonso {\sc Medina-Urrea}\inst{1} \and Juan-Manuel {\sc Torres-Moreno}\inst{2}$^,$\inst{3}
}

\institute{
  El Colegio de M\'exico,\\
  Carretera Picacho Ajusco 20, 14110 Tlalpan CDMX, México\\
  \email{amedinau@colmex.mx}
\and
    Polytechnique Montr\'eal,\\ CP. 6128 succursale Centre-ville,
   Montr\'eal (Québec) Canada\\
\and
  Laboratoire Informatique d'Avignon,\\
  BP 91228 84911, 
  Avignon, Cedex 09, France\\
  \email{juan-manuel.torres@univ-avignon.fr}
 }

\authorrunning{\sc Medina-Urrea and Torres-Moreno}

\titlerunning{RIMAX: Ranking Semantic Rhymes}

\begin{document}

\maketitle

\begin{abstract}
This paper presents RIMAX, a new system for detecting semantic rhymes, using a Comprehensive Mexican Spanish Dictionary (DEM) and its Rhyming Dictionary (REM).
We use the Vector Space Model to calculate the similarity of the definition of a query with the definitions corresponding to the assonant and consonant rhymes of the query.
The preliminary results using a manual evaluation are very encouraging.
\end{abstract}



%
%

\section{Introduction}
\label{sec:intro}

Given that a rhyme is a set of two or more words sharing an ending, a semantic rhyme is a set of words sharing an ending and exhibiting similar meanings, i.e. they are semantically related. 

Our main motivation for developing a system for detecting semantic rhymes is that, to the best of our knowledge, this kind of system does not exist. 
In order to create such a system, we need minimally some ingredients: 1) a rhyming dictionary, 2) the set of definitions of those rhymes and 3) a strategy to measure semantic proximity. 

We think this approach can be applied to different romance languages, like Spanish, French or Portuguese. 
Nevertheless, who have chosen the Spanish language spoken in Mexico, given the availability of its resources, like the Dictionary of Mexican Spanish (DEM)\footnote{\textit{Diccionario del español de M\'exico}, \url{https://dem.colmex.mx/}.} \cite{dem} and its rhyming dictionary (REM) \cite{rem:18}.

Rhyming dictionaries gather words according to rhyming patterns. 
In Spanish, rhymes can be either \textit{consonant} or \textit{assonant}. The former kind of rhyme shares ending sequences of vocalic and consonant sounds; the latter shares similar vowel sounds. In other words, they are based on pronunciation, not on writing patterns. 

Also, since consonance and assonance depend on the stressed syllable, words which end with a stressed syllable are grouped together, those whose stressed syllable is the next to last appear together, and so on.  
For example, the penultimate syllables of the following Spanish words are the stressed syllables: \textit{bula, gula, angula, chula, mula, chamula}. So these words should appear together in a rhyming dictionary.

It is very important to note that word pronunciation may vary with time and across geographical and social dialects. 
In Spanish, this is particularly evident when word loans are considered (for instance, Anglicisms and Gallicisms), because they tend to keep their original writing. 
For example, the following rhymes are loan words and they are common in Mexican Spanish: \textit{flash, collage, garage, cottage, squash}. 
Their last syllable is stressed and they are ordered in reverse according to their sounds and not their letters (respectively, /\textipa{"flaS}/, /\textipa{ko."laS}/, /\textipa{ga."raS}/, /\textipa{ko."taS}/ and /\textipa{es."kwaS}/).

This paper is organized as follows.
The rhyming dictionary is presented in Section \ref{sec:diccRimas}.
Section \ref{sec:method} presents our strategies for ranking rhymes.
Evaluation is presented in Section \ref{sec:evaluation}, followed by a discussion and final remarks in Section \ref{sec:conclusion}.





\section{A Rhyming Dictionary}
\label{sec:diccRimas}

In this experiment, we used the nomenclature of the dictionary of Mexican Spanish (DEM).
The nomenclature is the set of words, in their cannonical form, defined in a dictionary. 
So the DEM's nomenclature was used to generate automatically a phonological transcription. 
This is needed for classifying words according to their stressed syllable. In a rhyming dictionary, Spanish words are generally grouped into three sets:  

\begin{enumerate}
 \item \textbf{Oxytone words}. 
 Their last syllable is stressed (e.g. Sp: \textit{color} /\textipa{ko."lor}/; Fr: \textit{éboche} /\textipa{e."boS}/; En: \textit{correct} /\textipa{k@."rEkt}/). 
 
 \item \textbf{Paroxytone words}. 
 Their second to last (penultimate) syllable is stressed (e.g. Sp: \textit{mano} /\textipa{"ma.no}/; En: \textit{second} /\textipa{"sE.k@nd}/). 
 
 \item \textbf{Proparoxytone words}. 
 Their third to last (antipenultimate) syllable is stressed (e.g. Sp: \textit{br\'ujula} /\textipa{"bru.hu.la}/; En; \textit{conversational} /\textipa{k6n.ve\textrhoticity."seI.S@n.@l}/).  
\end{enumerate}

So, typically, a Spanish rhyming dictionary separates these three types of words\footnote{This is not always the case. For instance, see Bloise Campoy’s \cite{bloise:46}, which groups words in five sets.}. A set of rhyming words belongs to a subset of these groups. 
As mentioned above, there are at least two kinds of rhymes relevant to this project:

\begin{itemize}
 \item \textbf{Consonant Rhyme}. 
Two or more words are in consonance if they share exactly all sounds from their stressed vowel to the end, e.g. Sp: \textit{br\textbf{újula}} /\textipa{\textbf{"}br\textbf{u.hu.la}}/ and \textit{esdr\textbf{újula}} /\textipa{es.\textbf{"}dr\textbf{u.hu.la}}/; both belong to the set of words ending with $\sim$\textit{\'ujula}.
 
 \item \textbf{Assonant Rhyme}. 
Two or more words are in assonance if they share all vocalic sounds from their stressed syllable to the end, e.g., Sp: \textit{br\textbf{ú}j\textbf{u}l\textbf{a}} /\textipa{\textbf{"}br\textbf{u}.x\textbf{u}.l\textbf{a}}/ and \textit{p\textbf{ú}rp\textbf{u}r\textbf{a}} /\textipa{\textbf{"}p\textbf{u}r.p\textbf{u}.r\textbf{a}}/ (both belong to the set of words ending with the vocalic pattern *.\'u.u.a).
\end{itemize}

Let $R$ be a rhyming dictionary composed of sets $A$, $G$ and $E$ which contain, respectively, oxytone, paroxytone and proparoxytone words. 
Each of these sets contains sets of words sharing vocalic patterns. 
Thus, the set of oxytones whose last syllable contains /\'a/ can be named \textbf{*.\'a}. 
Those whose last syllable containss /\'e/ can be named \textbf{*.\'e} and so on. 
The set of paroxytones whose stressed syllable contains /\'a/ and last syllable contains /a/ can be named \textbf{*.\'a.a}.
Those whose stressed syllable contains /\'a/ and last syllable contains /e/ can be named \textbf{*.\'a.e} and so fort.
Finally, the set of proparoxytones whose stressed syllable contains /\'a/, the penultimate syllable contains /e/ and last syllable contains /a/ can be named \textbf{*.\'a.e.a}.
Those whose stressed syllable contains /\'a/, the penultimate syllable contains /e/ and last syllable contains /e/ can be named \textbf{*.\'a.e.e} and so on and so fort.

Each vocalic pattern set contains assonant rhymes. For example, both \textit{br\'ujula} and \textit{p\'urpura} belong to set \textbf{*.\'u.u.a}.

Furthermore, within these assonant rhyme sets, consonant rhyme sets can be found. 
For instance, all words ending with $\sim$\textit{\'abana} (\textit{guan\'abana}, \textit{s\'abana}), $\sim$\textit{ámara} (\textit{c\'amara}, \textit{rec\'amara}, \textit{antec\'amara}, \textit{s\'amara}), $\sim$\textit{\'ascara} (\textit{c\'ascara}, \textit{m\'ascara}) constitute subsets of \textbf{*.\'a.a.a}.
Thus, word \textit{c\'amara} belons to set $\sim$\textit{ámara} and $\sim$\textit{\'amara} is a subset of \textbf{*.\'a.a.a}
(\textit{c\'amara} $\in \sim$\textit{\'amara}, $\sim$\textit{\'amara} $\subset$ \textbf{*.\'a.a.a}).
More importantly, \textit{c\'amara} and \textit{s\'amara} are consonant rhymes because they belong to the same set $\sim$\textit{\'amara}.

In summary, a rhyming dictionary of this sort can be generated automatically by transcribing phonologically the desired lexicon and ordering it accordingly, e.g. the DEM's nomenclature, in order to separate words according to their stressed syllable and group them according to the their ending.

\section{Rhyme Ranking by Definition Similarity}
\label{sec:method}

Online dictionaries can do much more than recover and display information.
Even if their aim is merely to display lists of words, like reverse or rhyming dictionaries do, they offer such downright, simple advantages as ranking and ordering of results.

From a language perspective, it is interesting that text mining techniques can be applied to accomplish this. 
In fact, text similarity measures can be used to find out how similar word definitions are, i.e. measuring definition similarity.  
 
Let $D$ be a dictionary containing the set of defined words $w$ and the set of definitions $d$. 
Since a word may have several senses, let $d_{ij}$ be the $j$th definition of word $w_i$ in $D$. 
Similarly, let $d'_{kl}$ be the $l$th definition of word $w_k$. 
Also, let $v_{ij}$ and $v'_{kl}$ be vectors where the frequencies of lemmatized or ultrastemmized \cite{DBLP:journals/corr/abs-1209-3126} content words of definitions $d_{ij}$ and $d'_{kl}$ are stored.

Then, the similarity between $d_{ij}$ and $d'_{kl}$ can then be measured using well-known techniques \cite{manning:99,manning:08}:

\begin{itemize}
 \item Cosine Similarity Measure $s_c$
 \item Dice Similarity Coefficient $s_d$
 \item Euclidean Distance $s_e$
 \item Manhattan Distance $s_m$
 \item Levenshtein Distance $s_l$
 \item Jaccard Distance $s_j$
\end{itemize}

In order to find semantic rhymes, each member of the rhyming set of word $x$ will be weighed according to how similar its definition is to that of $x$, using any of these measurement strategies. 
Hence, given a query word, consonance and assonance lists are generated and ordered by the calculated similarity among definitions. 

\section{Rimax}
\label{sec:rimax}

RIMAX (\textit{Rimes MAXimales}) is a system that finds semantically associated consonant and assonant rhymes of Mexican Spanish.
RIMAX is written in PERL 5.0, because it allows using easily regular expressions to map the rhyming patterns with the query.

As mentioned before, RIMAX uses the following resources:

\begin{itemize}
\item	A Comprehensive Dictionary of Mexican Spanish (DEM) 
\item	A Dictionary  of rhymes of Mexican Spanish (REM) 
\item	A Lemmatization Dictionary of the same language (from the Cortex summarization system) \cite{ats:14}.
\end{itemize}

The program allows to select the functions of similarity and a suitable threshold (\textit{seuil}) for each experiment and it saves them in a control file
that is referenced in subsequent experiments. 

The similarity measures mentioned in section \ref{sec:method} were all implemented in RIMAX, but we use only the cosine similarity measure and a threshold = 0 for the evaluation below.


\section{Evaluation}
\label{sec:evaluation}

As a preliminary evaluation of RIMAX, we asked a set of 6 individuals (students from various educational levels, native speakers of Spanish from Mexico) to answer a questionnaire of 10 questions.
Each question offered three sets of rhymes to be selected by the individual. 
They were instructed to pick the best set rhymes, given a requested word. 
That is, each question includes one word and three sets of its rhymes.

The three sets of words are: 1) a set of assonant rhymes generated by the RIMAX program, 2) another set of consonant rhymes also generated by RIMAX and 3) a set of rhymes as found in the pages of the paper dictionary \cite{rem:18}.
In each question, the sets appear in a random order (hidden from the individual).

For instance, for the word “tequila”, the question included the following sets of rhymes (the last of which is the set of assonants):

\vspace{0.2cm}

\noindent {\it ¿Qué conjunto de rimas se relaciona mejor con la palabra “tequila”?\footnote{What set of rhymes best relates to the word “tequila”?}

\begin{description}
\item a) ventila, anguila, maquila, esquila, axila

\item b) pila, güila, redila, gorila, clorofila
 
\item c) sidra, sangrita, cantina, trementina, bebida 
\end{description}
}

Although there are many words that do not have consonant rhymes, i.e. they are dissonant words, we decided not to take them into account in this evaluation to guarantee that we always have three different groups of rhymes for each question. 

The list of the 10 words used for the questions of the questionnaire is the following one:

\vspace{0.2cm}
 
\noindent Q=\{ {\it tequila, tortilla, camarón, pelota, comadre, corazón, jalapeño, vampiro, computadora, mariguana} \}

\vspace{0.2cm}

Table \ref{tab:exp} shows the results obtained in our evaluation experiment. Each of the three columns corresponds to the number of individuals who voted for each of the types of rhyme sets in every question.

\begin{table}[h]
\centering
	\caption{Evaluation experiment}
	\label{tab:exp}
	\begin{tabular}{|c|c|c|c|c|}
	\hline
          & Consonant & Assonant & Dictionary \\
 Question & Rhymes    & Rhymes   & Rhymes \\
\hline
1)	&0	&4	&2\\
2)	&2	&2	&2\\
3)	&3	&0	&3\\
4)	&2	&2	&2\\
5)	&6	&0	&0\\
6)	&4	&0	&2\\
7)	&2	&2	&2\\
8)	&2	&3	&1\\
9)	&4	&0	&2\\
10)	&0	&4	&2\\
	\hline
	\end{tabular}
\end{table}

As can be seen in these evaluations, 42\% of the individuals chose the consonant rhymes generated by the program.
28\% the assonant rhymes generated by the program and 30\% the dictionary rhymes.
In other words, 70 percent prefer RIMAX-generated rhymes to dictionary rhymes.

\section{Discussion and Final Remarks}
\label{sec:conclusion}

We anticipated that individuals would pick more the assonant rhymes over the consonant ones, because there is a larger set of definitions from which their semantic ranking would be calculated.
However, the consonant rhymes were the most picked, probably because the assonant rhymes look less like a proper rhyme. 

Some final remarks have to do with the general performance of RIMAX. We can say, for example, that some rhymes are very numerous and require a lot of processing. 
One way of dealing with this is storing similarity values in a matrix so that future processes can be faster. 


\bibliographystyle{plain}
\bibliography{biblio}

\begin{thebibliography}{1}

\bibitem{bloise:46}
Pascual Bloise~Campoy.
\newblock {\em {Diccionario de la rima}}.
\newblock Aguilar, Madrid, 1946.

\bibitem{dem}
Luis~Fernando Lara~(coord).
\newblock {\em {Diccionario del Espa\~nol de México}}.
\newblock El Colegio de México, México, 2010.

\bibitem{manning:08}
Christopher~D. Manning, Prabhakar Raghavan, and Hinrich Schütze.
\newblock {\em {Introduction to Information Retrieval}}.
\newblock Cambridge University Press, 2008.

\bibitem{manning:99}
Christopher~D. Manning and Hinrich Sch{\"u}tze.
\newblock {\em {Foundations of Statistical Natural Language Processing}}.
\newblock MIT Press, Cambridge, 1999.

\bibitem{rem:18}
Alfonso Medina~Urrea.
\newblock {\em {Diccionario de rimas asonantes y consonantes del español de
  México}}.
\newblock El Colegio de México, Mexico, 2018.

\bibitem{DBLP:journals/corr/abs-1209-3126}
Juan{-}Manuel Torres{-}Moreno.
\newblock Beyond stemming and lemmatization: Ultra-stemming to improve
  automatic text summarization.
\newblock {\em CoRR}, abs/1209.3126, 2012.

\bibitem{ats:14}
Juan-Manuel Torres-Moreno.
\newblock {\em {Automatic Text Summarization}}.
\newblock Wiley-ISTE, London, 2014.

\end{thebibliography}

\end{document}